\documentclass[preprint]{article}

\usepackage{arxiv}

\usepackage[utf8]{inputenc} 
\usepackage[T1]{fontenc}    
\usepackage{hyperref}       
\usepackage{url}            
\usepackage{booktabs}       
\usepackage{amsfonts}       
\usepackage{nicefrac}       
\usepackage{microtype}      
\usepackage{lipsum}
\usepackage{graphicx}
\usepackage{framed,multirow}
\usepackage{amsmath,amssymb,amsthm}
\usepackage{algorithmic}
\usepackage[ruled]{algorithm}
\usepackage{upgreek}
\usepackage{latexsym}
\usepackage{url}
\usepackage{xcolor}
\usepackage{hyperref}
\usepackage{booktabs}
\usepackage{parskip}
\usepackage{subfiles}
\newcommand{\B}{\bfseries}

\usepackage{array}
    \newcolumntype{P}[1]{>{\centering\arraybackslash}p{#1}}
    \newcolumntype{M}[1]{>{\centering\arraybackslash}m{#1}}

\newsavebox\CBox
\def\textBF#1{\sbox\CBox{#1}\resizebox{\wd\CBox}{\ht\CBox}{\textbf{#1}}}

\definecolor{myred}{rgb}{.85, 0, 0}

\newcommand{\first}[1]{\textcolor{myred}
{\textBF{#1}}} 
\newcommand{\second}[1]{\textcolor{blue}
{\textBF{#1}}}

\newcommand\figref{Fig.~\ref}
\newcommand\tabref{Table~\ref}

\newcommand{\beq}{\begin{equation}}
\newcommand{\eeq}{\end{equation}}

\newcommand{\vect}[1]{\mathbf{#1}}

\newcommand{\mr}[1]{\mathrm{#1}}

\newcommand{\CaVAT}{C\textsc{a}VAT}

\newcommand{\ff}{\vect{f}}
\newcommand{\pp}{\vect{p}}

\DeclareMathOperator*{\argmax}{arg\,max}

\newcommand{\img}{x}

\newcommand{\gt}{y}
\newcommand{\lab}{\mathcal{S}}
\newcommand{\unlab}{\mathcal{U}}
\newcommand{\data}{\mathcal{D}}
\newcommand{\voxels}{\Upomega}
\newcommand{\params}{\uptheta}
\newcommand{\classes}{\mathcal{C}}
\newcommand{\loss}{\mathcal{L}}

\newcommand{\kl}[2]{D_{\mr{KL}}\big(#1 \, || \, #2\big)}
\newcommand{\pseudol}{\widehat{y}}
\DeclareMathOperator{\expect}{\mathbb{E}}

\newcommand{\mypar}[1]{\noindent\textbf{#1}~~}

\newcommand{\omitme}[1]{}

\newcommand{\myvspace}[1]{}

\newcommand{\visComp}[1]{\includegraphics[width=0.11\textwidth]{Figures/#1.png}}

\title{Context-aware virtual adversarial training for anatomically-plausible segmentation}

\author{
  Ping Wang\thanks{Corresponding author} \\
  Department of Software and IT Engineering\\ Ecole de technologie sup\'erieure\\
  Montreal, H3C1K3, Canada \\
  \texttt{ping.wang.1@ens.etsmtl.ca} \\
   \And
 Jizong Peng \\
  Department of Software and IT Engineering\\ Ecole de technologie sup\'erieure\\
  Montreal, H3C1K3, Canada \\
  \texttt{jizong.peng.1@etsmtl.net} \\
  \And
  Marco Pedersoli \\
  Department of Systems Engineering\\ 
  Ecole de technologie sup\'erieure\\
  Montreal, H3C1K3, Canada \\
  \texttt{marco.pedersoli@etsmtl.ca} \\
  \And
  Yuanfeng Zhou \\
  School of Computer Science and Technology\\
  Shandong University\\ 
  Jinan, 250101, China\\
  \texttt{yfzhou@sdu.edu.cn} \\
  \And
  Caiming Zhang \\
  School of Computer Science and Technology\\
  Shandong University\\ 
  Jinan, 250101, China\\
  \texttt{czhang@sdu.edu.cn} \\
  \And
  Christian Desrosiers \\
  Department of Software and IT Engineering\\ 
  Ecole de technologie sup\'erieure\\
  Montreal, H3C1K3, Canada \\
  \texttt{christian.desrosiers@etsmtl.ca} \\
}

\begin{document}
\maketitle

\begin{abstract}
Despite their outstanding accuracy, semi-supervised segmentation methods based on deep neural networks can still yield predictions that are considered anatomically impossible by clinicians, for instance, containing holes or disconnected regions. To solve this problem, we present a Context-aware Virtual Adversarial Training (\CaVAT) method for generating anatomically plausible segmentation. Unlike approaches focusing solely on accuracy, our method also considers complex topological constraints like connectivity which cannot be easily modeled in a differentiable loss function. We use adversarial training to generate examples violating the constraints, so the network can learn to avoid making such incorrect predictions on new examples, and employ the R\textsc{einforce} algorithm to handle non-differentiable segmentation constraints. The proposed method offers a generic and efficient way to add any constraint on top of any segmentation network. Experiments on two clinically-relevant datasets show our method to produce segmentations that are both accurate and anatomically-plausible in terms of region connectivity.
\end{abstract}

\section{Introduction} 

Due to the high complexity and cost of generating ground-truth annotations for medical image segmentation, a wide range of semi-supervised methods based on deep neural networks have been proposed for this problem. These methods, which leverage unlabeled data to improve performance, include distillation~\citep{radosavovic2018data}, attention learning~\citep{min2018robust}, adversarial learning~\citep{souly2017semi,zhang2017deep}, entropy minimization~\citep{vu2019advent}, co-training~\citep{peng2020deep,zhou2019semi}, temporal ensembling~\citep{perone2018deep,cui2019semi}, consistency-based regularization~\citep{bortsova2019semi} and data augmentation~\citep{chaitanya2019semi,zhao2019data}. When very few labeled images are available, however, it may be impossible for a segmentation network to learn the distribution of valid shapes, even when using a semi-supervised learning approach. As a result, the segmentation network can yield predictions that are considered anatomically impossible by clinicians \citep{painchaud2020cardiac}. Such predictions can severely impact downstream analyses which rely on anatomical measures, and often require a costly manual step to correct segmentation errors. 

Various works have focused on incorporating constraints in semi-super\-vised or weakly-supervised segmentation methods \citep{kervadec2019constrained,pathak2015constrained,jia2017constrained,zhou2019prior,Masoud2016}. The approach in \citep{jia2017constrained} uses a simple $L_2$ penalty to impose size constraints on segmented regions in histopathology images. Kervadec et al. \citep{kervadec2019constrained} proposed a similar differential loss to enforce inequality constraints on the size of segmented regions. Likewise, Zhou et al. \citep{zhou2019prior} constrain the size of segmented regions with a loss function minimizing the KL divergence between the predicted class distribution and a target one. Despite showing the benefit of adding constraints in a segmentation model, these methods suffer from two important limitations. First, they are limited to simple constraints like region size or centroid position, which are insufficient to characterize the complex shapes found in medical imaging applications. Second, they require designing a problem-specific differentiable loss and, thus, have low generalizability. 

Recent efforts have also been invested toward adding strong anatomical priors in segmentation networks. In \citep{oktay2017anatomically}, Oktay et al. present an anatomically constrained neural network (ACNN) using an autoencoder to reconstruct the segmentation mask of labeled images. The reconstruction loss of the autoencoder for a given image is then used as segmentation shape prior. As training the autoencoder requires a sufficient amount of labeled data, this approach is poorly suited to semi-supervised learning settings. The cardiac segmentation approach by Zotti et al. \citep{zotti2018convolutional} improves accuracy by aligning a probabilistic shape atlas to the predicted segmentation during training. Likewise, Duan et al. \citep{duan2019automatic} uses a multi-task approach to locate landmarks which guide an atlas-based label propagation during a refinement step. In spite of their added robustness, both theses approaches need large annotated datasets to learn the atlas and are sensitive to atlas registration errors. Recently, Painchaud et al. \citep{painchaud2020cardiac} proposed a segmentation method that uses a variational autoencoder to learn the manifold of valid segmentations. During inference, predicted segmentations are mapped to their nearest valid point in the manifold. While it offers strong anatomical guarantees, this post-processing method requires pre-computing an important number of valid points. Moreover, the projection of a predicted output on these points can lead to a segmentation considerably different from the ground-truth.

To address the above-mentioned limitations, we propose a Context-Aware Virtual Adversarial Training (\CaVAT) method for semi-supervised segmentation, which considers complex constraints during training to learn an anatomically-plausible segmentation. Unlike existing approaches, which are limited to simple, differentiable constraints (e.g., region size, centroid position, etc.) and require designing a customized loss function, our method can be used out-of-the-box to add any constraint, differentiable or not, on top of a given segmentation model. Our detailed contributions are as follows:
\begin{itemize}
\item We propose a novel framework that helps obtain anatomically-plausible segmentations by considering complex anatomical priors in the learning process. Our framework is based on Virtual Adversarial Training (VAT)~\citep{Miyato2019}, which optimizes a minimax problem where adversarial examples are created from training samples so to maximize prediction divergence of the network. Unlike VAT, our method generates adversarial examples that maximize prediction divergence \emph{as well as} constraint violation. The R\textsc{einforce} algorithm~\citep{Williams1992} is used to compute gradients for non-differentiable segmentation constraints.
\item To our knowledge, our segmentation method is the first to consider complex anatomical priors in a general semi-supervised setting. In comparison, existing approaches require a large number of labeled images to learn a shape prior \citep{oktay2017anatomically,painchaud2020cardiac} or a complex and problem-specific step involving atlas registration \citep{duan2019automatic,dong2020deep}. Unlike these approaches, our method needs very few labeled examples and can be used with any segmentation network.
\end{itemize}
In the next section, we present our Context-aware Virtual Adversarial Training (\CaVAT) method and show how it can be used to include connectivity constraints on the segmentation output. In our experiments, we demonstrate our semi-supervised segmentation method's ability to provide a higher accuracy and better constraint satisfaction when trained with very few labeled examples. Finally, we conclude with a summary of main contributions and results.

\section{Proposed method} 

We start by defining the semi-supervised segmentation problem considered in our work. Let $\lab=\{(\img_s,\gt_s)\}_{s=1}^{|\lab|}$ be a small set of labeled examples, where each $\img_s \in \mathbb{R}^{|\voxels|}$ is an image and $\gt_s \in \{0,1\}^{|\voxels|\times|\classes|}$ is the corresponding ground-truth segmentation mask. Here, $\voxels \subset \mathbb{Z}^2$ denotes the set of image pixels and $\classes$ the set of segmentation classes. Given labeled images $\lab$ and a larger set of unlabeled images $\unlab=\{\img_u\}_{u=1}^{|\unlab|}$, we want to learn a network $f$ parameterized by weights $\params$ which produces segmentations that are both accurate and anatomically-plausible. 

\begin{figure}
    \centering
    \includegraphics[width=1\textwidth]{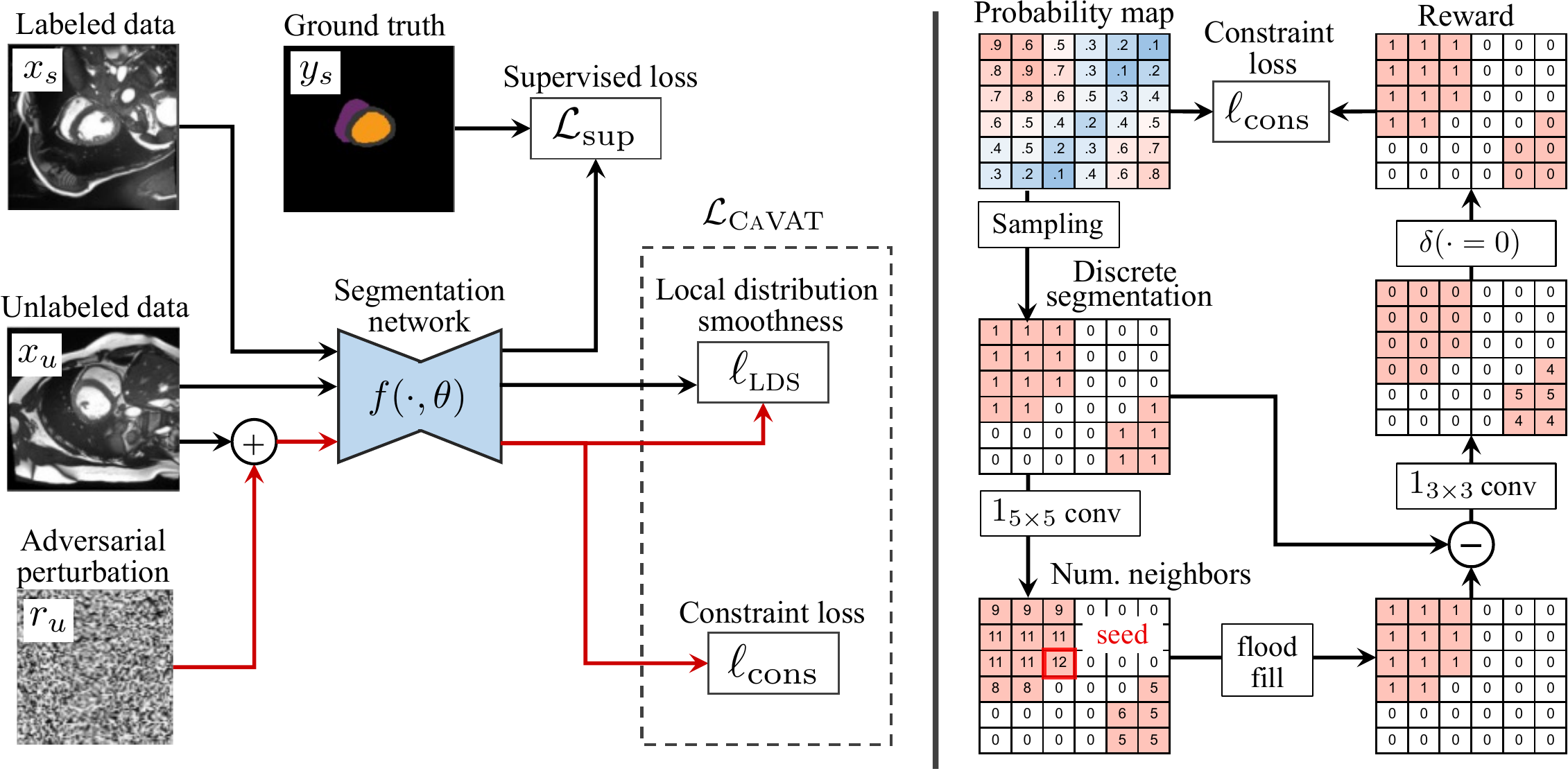}
    \vspace{-5mm}
    \caption{Overview of the proposed method.}
    \label{fig:overview}
    \myvspace{-3mm}
\end{figure}

An overview of the proposed method is shown in Fig.~\ref{fig:overview} (left). Our method is trained with both labeled and unlabeled data by optimizing the following objective:
\beq
\label{eq:total_loss}
 \min_{\theta} \ \loss_{\mr{total}}(\params; \, \data) \ = \ \loss_{\mr{sup}}(\params;\lab) \, + \, \lambda\loss_{\mr{\CaVAT}}(\params;\unlab)
\eeq 
The supervised loss $\loss_{\mr{sup}}(\cdot)$ encourages individual networks to predict segmentation outputs for labeled data that are close to the ground truth. In this work, we use the well-know cross-entropy loss:
\beq
 \loss_{\mr{sup}}(\params;\,\lab) \ = \ -\frac{1}{|\lab|} \sum_{(\img, \gt)\in\lab}\sum_{i \in \voxels}\sum_{j\in\classes} y_{ij} \log\big(\ff_{ij}(\img,\params)\big)
\eeq 
The context-aware VAT loss $\loss_{\mr{\CaVAT}}(\cdot)$, which uses unlabeled images, increases the robustness of the model to adversarial noise and helps the model learn to produce valid segmentations with respect to the given constraints. This loss is detailed in the next section.

\subsection{Context-aware VAT loss}

A standard approach to incorporate constraints in a semi-supervised learning scenario is to add a loss term that penalizes the violation of these constraints~\citep{kervadec2019constrained,jia2017constrained,zhou2019prior}. This simple approach poses three major problems. First, it may not be possible to model a given constraint with a function. For instance, testing region connectivity, which imposes each pair of points in a region to be connected by a path inside the region, requires running an algorithm. Second, even if such function exists, it may not be differentiable. This is often the case in segmentation due to the discrepancy between the continuous network output and the discrete segmentation on which the constraints are applied. Last, although both these conditions are satisfied, there is no guarantee that a given constraint will be violated during training, especially if it models a complex relationship. If the network never violates a constraint, it will not be able to learn how to satisfy it since the gradient from the constraint loss will be null.     

To alleviate these problems, we define the following Context-aware VAT loss on unlabeled examples:
\beq
\label{eq:CaVAT_general}
 \loss_{\mr{\CaVAT}}(\params; \unlab) \ = \
 \frac{1}{|\unlab|}\sum_{x_u \in \unlab} \,\max_{\|r\|\leq \epsilon} \Big[ \ell_{\mr{\textsc{lds}}}(x_u, r_u; \params) \, + \gamma \, \ell_{\mr{cons}}(x_u, r_u; \params) \Big]
\eeq
This loss, which is composed of a local distribution smoothness (LDS) term $\ell_{\mr{\textsc{lds}}}$ and reinforced constraint term $\ell_{\mr{cons}}$ is minimized with respect to network parameters $\params$ and maximized with respect to the image perturbation $r_u$. $\gamma$ is the weight balancing the two loss terms, which are described below. \\[-3pt] 

\mypar{Local distributional smoothness (LDS)} The first term in Eq.~(\ref{eq:CaVAT_general}) is the divergence-based LDS in the original VAT method~\cite{Miyato2019} which is given by
\begin{equation}\label{eq:vat}
\ell_{\mr{\textsc{lds}}}(x_u, r_u; \params) \ = \ \kl{\ff(x_u;\,\params)}{\ff(x_u+r_u;\,\params)} 
\end{equation} 
Minimizing $\ell_{\mr{\textsc{lds}}}(\cdot)$ enhances the robustness of the model against adversarial examples that violates the virtual adversarial direction, thereby improving generalization performance.\\[-2pt] 

\mypar{Reinforced constraint loss}
The second term in Eq.~(\ref{eq:CaVAT_general}) encourages the production of adversarial examples leading to violated constraints, which is necessary for learning these constraints. The reinforced constraint loss is given by
\beq\label{eq:policy_gradient}
\ell_{\mr{cons}}(x_u,r_u; \params)\ = \ 
-\expect_{\,\pseudol \sim \ff(\img_u + r_u; \params)} \Big[ J(\pseudol)\Big]
\eeq
where $\pseudol$ is discrete segmentation mask sampled from the output probability distribution $\pp_u =\ff(\img_u\!+\!r_u; \params)$ for an adversarial image $\img_u+r_u$, and $J$ is the reward function which outputs 1 if the constraint is satisfied else it returns 0. Since the discrete segmentation sampling step is non-differentiable, we resort to the R\textsc{einforce} algorithm~\citep{Williams1992} to convert it into a differentiable loss:
\beq
\nabla_{\params}\ell_{\mr{cons}} \ = \ -\sum_{\pseudol} J(\pseudol)\nabla_{\params} p(\pseudol)
  \ \approx \ -\frac{1}{m} \sum_{\pseudol^{(s)}\sim \pp_u, \, s=1}^{m}\!\!\!\!\!\! J(\pseudol^{(s)})\,\nabla_{\params} \log p\big(\pseudol^{(s)}\big)
\eeq
where $m$ is a given number of samples, empirically set to 10 in this paper. Assuming that outputs at different pixels are conditionally independent given the input image, i.e. $p(\pseudol^{(s)}) = \prod_{i}p\big(\pseudol_i^{(s)})$, the final loss can be expressed as
\beq\label{eq:policy_gradient}
\ell_{\mr{cons}}(x_u,r_u; \params)\ = \ -\frac{1}{m} \sum_{\pseudol^{(s)}\sim \pp_u, \, s=1}^{m}\sum_{i\,\in\,\voxels}J_i(\pseudol^{(s)})\, \log p\big(\pseudol_i^{(s)}\big).
\eeq

\subsection{Local connectivity constraints}\label{sec:local-connectivity}

Although our method can be used with any differentiable or non-differentiable constraint, in this paper, we illustrate it on a well-known constraint with broad applicability: connectivity. Given a segmented region $G$, we say that $G$ is connected if and only if there exists a path between each pair of pixels $p,q \in G$ such that all pixels in the path belong to $G$. 
Imposing connectivity in segmentation leads to a highly-complex problem which can only be solved for simplified cases, for example, by representing an image as a small set of superpixels \citep{shen2020ilp}. However, considering connectivity over the whole image may not be practical since it is hard to achieve in the early training stages. For instance, having a single disconnected noisy pixel violates the constraint. To solve this problem, we relax the global constraint and instead consider connectivity at each local patch. Since satisfaction at local patches is a necessary condition for global satisfaction, enforcing it helps achieve our objective. Moreover, doing so provides a spatially-denser gradient since satisfaction can vary from one sub-region to another.

The reward computation process is illustrated in Fig 1 (right) and detailed in Algorithm 1 of the Supplementary Materials. First, we generate discrete segmentations $\pseudol^{(s)}$, $s=1,\ldots,m$, from the output probability map via multinomial sampling. For each sampled $\pseudol^{(s)}$, we then apply the flood-fill algorithm from a chosen seed pixel to produce the connected foreground region $C^{(s)}$. To select the seed pixel, we use a $\text{1}_{l \times l}$ convolution kernel on $\pseudol^{(s)}$ to compute the number of foreground pixels in a $l\!\times\!l$ window centered on each pixel of the image. Afterwards, we randomly choose a pixel with maximum value to favor selecting large connected components as reference region. For each patch of $k\!\times\!k$, we measure the number of foreground pixels that are not in $C^{(s)}$, using a simple convolution: $S^{(s)} = \text{1}_{k \times k} \circledast (\pseudol^{(s)}-C^{(s)})$. Finally, we evaluate the constraint at pixel $i$ as $J_i(\pseudol^{(s)})=\delta(S_i^{(s)} = 0)$, where $\delta(\cdot)$ is the Kronecker delta.

\begin{table}[!t]
\centering
\caption{Mean DSC, HD and non-connected pixels (N-conn) for segmenting the left ventricle (LV), right ventricle (RV) and myocardium (Myo) of ACDC, and segmenting prostate in P\textsc{romise}12. For each task, labeled data ratio and performance metric, we highlight the best method in \first{red} and the second best in \second{blue}.}
\label{tab:LV_bg}
\setlength{\tabcolsep}{6pt}
\renewcommand{\arraystretch}{.85}
\begin{scriptsize}
\begin{tabular}{llrrr}
\toprule
\B Task\,/\,labeled\,\% & \B Method&  \B DSC\,(\%)\,$\uparrow$ & \B HD\,(mm)\,$\downarrow$ &\B N-conn\,(\%)\,$\downarrow$\\ 
\midrule
ACDC LV\,/\,100\,\% & Baseline & 94.00 (0.09) & 6.02 (1.13) & 2.40 (0.72) \\ 
\midrule
\multirow{9}{*}{ACDC LV\,/\,3\,\%} & Baseline & 88.14 (0.67) & 20.58 (7.09) & 6.90 (0.81)\\
 & Entropy min & 87.54 (0.83) & 16.67 (0.75) & 6.49 (0.67)\\
 & VAT & 88.65 (0.45) & 20.66 (2.76) & 6.32 (0.88)\\
 & Co-training & 88.81 (0.39) & 12.36 (0.81) & 6.35 (0.28)\\
 & Mean Teacher & \second{90.91} (1.13) & \second{12.11} (1.97) & \second{3.36} (0.99)\\
\cmidrule(l{5pt}r{5pt}){2-5}
& \CaVAT\,($r_u\!=\!0$) & 88.63 (0.69) & 28.31 (5.78) & 6.89 (0.86)\\
 & \CaVAT & 89.18 (0.48) & 22.62 (0.45) & 4.75 (0.74)\\
 & CoT\,+\,\CaVAT & 90.21 (0.47) & 12.84 (3.56) & 5.94 (0.71)\\
 & MT\,+\,\CaVAT & \first{91.04} (0.60) & \first{9.52} (1.44) & \first{3.20} (0.75)\\
\specialrule{1pt}{2pt}{2pt}
ACDC Myo\,/\,100\,\% & Baseline &  89.55 (0.09) & 4.17 (0.19) & 1.91 (0.58)\\
\midrule
\multirow{9}{*}{ACDC Myo\,/\,3\,\%} & Baseline & 75.00 (2.55) & 27.85 (3.51) & 10.26 (2.23)\\
 & Entropy min & 74.01 (0.95) & 22.06 (3.90) & 11.68 (0.62)\\
 & VAT & 78.26 (0.62) & 26.45 (6.69) & 8.77 (0.77)\\
 & Co-training & 75.82 (0.39) & 13.24 (1.02) & 12.50 (0.71)\\
 & Mean Teacher & \second{82.56} (0.44) & \second{11.62} (1.80) & \second{4.26} (0.48)\\
\cmidrule(l{5pt}r{5pt}){2-5}
& \CaVAT\,($r_u\!=\!0$) & 78.44 (0.84) & 27.16 (1.05) & 6.48 (0.38)\\
 & \CaVAT & 79.59 (0.30) & 26.20 (0.59) & 6.52 (1.13)\\
& CoT\,+\,\CaVAT & 79.25 (1.03) & 12.34 (1.38) & 8.92 (0.22)\\
 & MT\,+\,\CaVAT & \first{82.68} (0.43) & \first{9.87} (0.74) & \first{3.82} (1.56)\\
\specialrule{1pt}{2pt}{2pt}
ACDC RV\,/\,100\,\% & Baseline & 88.66 (0.31) & 6.27 (0.38) & 6.32 (0.97)\\
\midrule
\multirow{9}{*}{ACDC RV\,/\,5\,\%} & Baseline & 63.17 (3.10) & 17.90 (0.87) & 27.50 (2.53)\\
 & Entropy min & 62.09 (1.22) & 16.72 (1.73) & 31.58 (2.70)\\
 & VAT & 69.52 (1.79) & 20.46 (3.69) & 25.81 (4.59)\\
 & Co-training & 63.97 (0.47) & 17.30 (1.58) & 29.07 (1.19)\\
 & Mean Teacher & \second{80.57} (0.65) & 
 \second{14.46} (1.81) & \second{12.21} (1.05)\\
\cmidrule(l{5pt}r{5pt}){2-5}
& \CaVAT\,($r_u\!=\!0$) & 70.42 (1.87) & 21.95 (2.75) & 21.52 (1.12)\\
 & \CaVAT & 72.88 (1.55) & 21.06 (3.42) & 20.43 (2.66)\\
 & CoT\,+\,\CaVAT & 71.51 (1.89) & 14.94 (2.04) & 25.92 (1.57)\\
 & MT\,+\,\CaVAT & \first{80.70} (0.51) & \first{11.90} (0.63) & \first{11.45} (1.19)\\
 \specialrule{1pt}{2pt}{2pt}
 \multirow{1}{*}{P\textsc{romise}12\,/\,100\,\%} & Baseline & 87.99 (0.20) & 5.04 (0.42) & 6.87 (0.19) \\
\midrule
\multirow{8}{*}{P\textsc{romise}12\,/\,5\,\%} & Baseline & 55.95 (1.80) & 11.86 (5.11) & 28.83 (2.18) \\
& Entropy min & 56.39 (3.01) & 10.95 (1.13) & 26.70 (1.88) \\
& VAT & 62.89 (4.20) & 14.12 (2.06) & 16.98 (4.31) \\
& Co-training & 52.60 (0.67) & 12.22 (2.91) & 34.60 (2.33) \\
& Mean Teacher & \second{71.09} (2.03) & \first{6.76} (3.92) & 16.19 (3.58) \\
\cmidrule(l{5pt}r{5pt}){2-5}
& \CaVAT\,($r_u\!=\!0$) & 63.68 (0.41) & 15.57 (0.58) & 15.12 (1.12)\\
& \CaVAT & 65.38 (2.24) & 14.55 (4.42) & \first{11.55} (0.43) \\
& CoT\,+\,\CaVAT & 66.65 (0.36) & 15.29 (0.28) & 11.57 (1.82) \\
& MT\,+\,\CaVAT & \first{72.33} (2.57) & \second{8.92} (0.06) & \second{12.33} (2.37) \\
\bottomrule
\end{tabular}
\end{scriptsize}
\vspace{-3mm}
\end{table}

\section{Experimental setup}

We evaluate our \CaVAT{} method on the Automated Cardiac Diagnosis Challenge (ACDC) dataset~\citep{ACDC} and the Prostate MR Image Segmentation (P\textsc{romise}12) Challenge dataset~\citep{litjens2014evaluation}. Details on these datasets can be found in the Supplementary Materials. For ACDC, segmentation masks delineate three anatomic regions: left ventricle endocardium (LV), left ventricle myocardium (Myo) and right ventricle endocardium (RV). All these regions satisfy the connectivity constraint and have a single connected component. For P\textsc{romise}12, the goal is to segment the whole prostate which is also a connected region. We report three performance metrics: Dice similarity coefficient (DSC), Hausdorff distance (HD) and Non-Connectivity (N-conn). DSC emphasises on the overall overlap between a candidate segmentation and its ground truth; HD measures the maximum local disagreement between the two segmentation sets; the N-conn quantifies the percentage of foreground pixels which are not connected to a randomly-selected foreground seed. The hyper-parameters for computing the connectivity reward (see Section \ref{sec:local-connectivity}) were set empirically as follows: $l=5$ and $k=3$. 

We tested labeled data ratios of $3\%$ and $5\%$ for each segmentation task,  and compared our \CaVAT{} method against  using only the supervised loss $\loss_{\mr{sup}}$ (denoted as Baseline in our results) as well as four popular approaches for semi-supervised learning: ~Entropy minimization~\citep{vu2019advent}, Virtual Adversarial Training (VAT)~\citep{Miyato2019}, Co-training~\citep{peng2020deep}, and Mean Teacher~\citep{cui2019semi}. Since our method can be used on top of any semi-supervised segmentation algorithm, we also evaluate its combination with Co-training (CoT\,+\,\CaVAT) or Mean Teacher (MT\,+\,\CaVAT). Last, we test our \CaVAT{} model with the same loss as in Eq.~(\ref{eq:CaVAT_general}) but no adversarial perturbation ($r_u=0$).

For all tested approaches, we use ENet~\citep{Paszke2016} as our segmentation backbone and train this network with a rectified Adam optimizer. The learning rate is initially set as to $1\times 10^{-5}$ and is updated by a warm-up and cosine decay strategy. We apply the same data augmentation as in \citep{peng2020discretely}. The hyper-parameter balancing the two terms of Eq. (\ref{eq:total_loss}) is set as follows: $\lambda=1\!\times\!10^{-3}$ LV, $\lambda=5\!\times\!10^{-4}$ for Myo, $\lambda=1\!\times\!10^{-4}$ for RV, and $\lambda=1\!\times\!10^{-4}$ for P\textsc{romise}12. For all experiments, we report the mean performance (standard deviation) on 3 independent runs with different random seeds.

\section{Experimental Results}
\myvspace{-2mm}
\tabref{tab:LV_bg} reports the DSC, HD and percentage of non-connected pixels (N-conn) on validation examples of the ACDC and P\textsc{romise}12 datasets. As can be seen, our \CaVAT{} method boosts performance in all cases compared to the baseline using only labeled images (Baseline), with DSC improvements of 1.04\% for ACDC LV, 4.59\% for ACDC Myo, 9.71\% for ACDC RV, and 9.43\% for P\textsc{romise}12. Our method also significantly reduces the number of non-connected foreground pixels (N-conn) compared to the baseline, demonstrating its ability to learn the given constraint. Results also validate the benefit of generating constraint-specific adversarial examples, as seen from the better DSC, HD and N-conn scores of \CaVAT{} compared to the setting with $r_u\!=\!0$. Moreover, we also observe improvements when adding \CaVAT{} to Co-training or Mean Teacher. In particular, our MT\,+\,\CaVAT{} combination obtains the highest overall DSC and yields a lower N-conn than Mean Teacher, for all segmentation tasks. Additional results with 5\% labeled data for ACDC LV and ACDC Myo, and with 8\% labeled data for P\textsc{romise}12 can be found in Supplementary Materials. 

In \figref{visualSEG}, we show examples of segmentations produced by the tested approaches for the three tasks, when using $5\%$ of labeled data. As can be seen, adding \CaVAT{} to the baseline or a semi-supervised learning method yields a more accurate segmentation and helps avoid disconnected regions. As last experiment, we performed a sensitivity analysis on hyper-parameter $\gamma$ which controls the weight of the constraint loss in Eq.~(\ref{eq:CaVAT_general}). The results and analysis for this experiment can be found in the Supplementary Materials. 

\begin{figure}[t!]
\centering
\renewcommand{\arraystretch}{1}
\setlength{\tabcolsep}{0.35pt}
\begin{footnotesize}
\begin{tabular}{ccccccccc}
\visComp{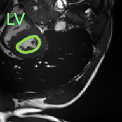} & 
\visComp{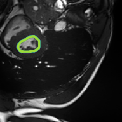} &
\visComp{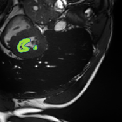} & 
\visComp{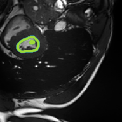} &
\visComp{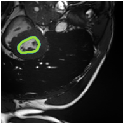} & 
\visComp{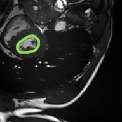} &
\visComp{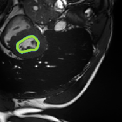} & 
\visComp{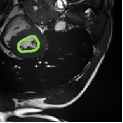} & 
\visComp{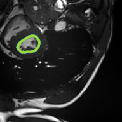}\\[-2pt]  
\visComp{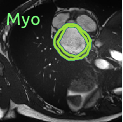} & 
\visComp{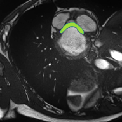} &
\visComp{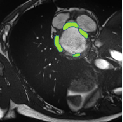} & 
\visComp{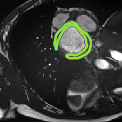} &
\visComp{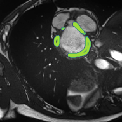} & 
\visComp{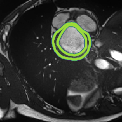} &
\visComp{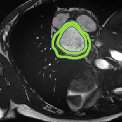} & 
\visComp{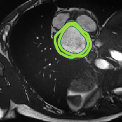} & 
\visComp{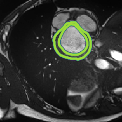}\\[-2pt]  
\visComp{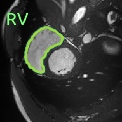} & 
\visComp{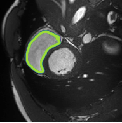} &
\visComp{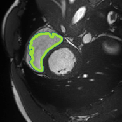} & 
\visComp{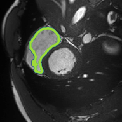} &
\visComp{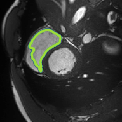} & 
\visComp{RV2_mt} &
\visComp{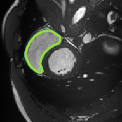} & 
\visComp{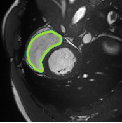} & 
\visComp{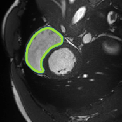}\\[-2pt] 
\visComp{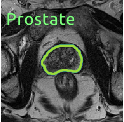} & 
\visComp{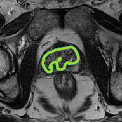} &
\visComp{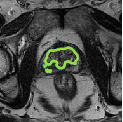} & 
\visComp{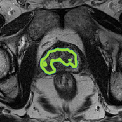} &
\visComp{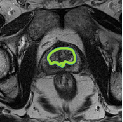} & 
\visComp{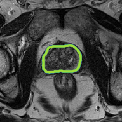} &
\visComp{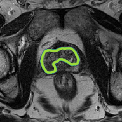} & 
\visComp{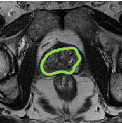} & 
\visComp{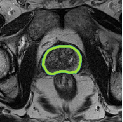}\\[-1pt] 
(a) & (b) & (c) & (d) & (e) & (f) & (g) & (h) & (i)
\end{tabular}
\end{footnotesize}
\myvspace{-3mm}
\caption{Visual comparison of tested methods on the validation images. (a) Ground-truth; 
(b) Partial supervision baseline; (c) Entropy min; (d) VAT; (e) Co-training (CoT); (f) Mean Teacher (MT); (g) \CaVAT; (h) CoT\,+\,\CaVAT; (i) MT\,+\,\CaVAT}\label{visualSEG}
\myvspace{-3mm}
\end{figure}

\section{Conclusion}

We proposed \CaVAT, a novel method for semi-supervised segmentation  that can incorporate complex anatomical constraints on any segmentation model during training. Our method extends the virtual adversarial training (VAT) framework, making a network robust to adversarial perturbations, by generating examples which also cause the model to violate a given constraint. By improving its prediction for these adversarial examples, the network can thus learn to satisfy the constraint. To alleviate the need to define a specialized penalty function for the constraint, as well as to handle non-differentiable constraints, our method uses the R\textsc{einforce} algorithm. As a result, it can be used as a plug-in on any semi-supervised learning approach. Experiments on three segmentation tasks from the ACDC and P\textsc{romise}12 datasets reveal the effectiveness of our method in terms of both the accuracy and constraint satisfaction.

A potential limitation of the proposed method stems from its use of the R\textsc{einforce} algorithm, which requires sampling a sufficient number of discrete segmentations from the predicted probabilities otherwise optimization may be unstable. While we found that 10 samples gave good results, a larger number might be required for more complex constraints. Another drawback of our method is the computational cost of evaluating constraints during training, which might be prohibitive in some cases. 
As future work, we plan to extend our method to multi-class segmentation tasks. We will also investigate the combination of our method with other semi-supervised techniques and evaluate its usefulness for a broader range of constraints.

\bibliographystyle{model2-names.bst}
\bibliography{Reference}

\vspace{8mm}
\textbf{\huge Supplementary Materials}\\

\author{}

\maketitle

\begin{scriptsize}
\begin{algorithm}[h]
Input: The segmentation $\pseudol^{(s)}$ sampling from output probability distribution of the model $\ff$, $s=1,\ldots,m$ \\
Output: Reward map $J_i(\pseudol^{(s)})$, $i\in \voxels$\\
\caption{Computation of the local connectivity reward}
\label{local_reward}
\emph{\textbf{Step 1}: Generating the connected foreground $C^{(s)}$}\\[.2em]
\setlength{\parindent}{.5em}Compute the number of foreground pixels $A^{(s)}_i$ in the patch around each pixel $i$ through convolution operation $A^{(s)} = \text{1}_{l \times l} \circledast \pseudol^{(s)}$\;\\
Randomly select a seed pixel $i^{\mr{seed}} \in \argmax_{i} \, A^{(s)}_i$\;\\
\mbox{Run the flood-fill algorithm from $i^{\mr{seed}}$ to get foreground connectivity map $C^{(s)}$\;}\\
\noindent\emph{\textbf{Step 2}: Estimating the reward map}\\[.2em]
\mbox{Get the map of non-connected pixels via convolution $S^{(s)} = \text{1}_{k \times k} \circledast (\pseudol^{(s)}-C^{(s)})$\;}\\
Compute the reward for pixel $i$ as $J_i(\pseudol^{(s)})=\delta(S_i^{(s)} = 0)$\;
\end{algorithm}
\end{scriptsize}

\textbf{Datasets}\\
\textbf{Automated Cardiac Diagnosis Challenge (ACDC):} This dataset consists of 200 MRI scans from 100 patients.
Scans correspond to end-diastolic (ED) and end-systolic (ES) phases, and were acquired on 1.5T and 3T systems with resolutions ranging from 0.70$\times$0.70\,mm to 1.92$\times$1.92\,mm in-plane and 5\,mm to 10\,mm through-plane. Three cardiac regions are labeled in the ground-truth: left ventricle (LV), right ventricle (RV) and myocardium (Myo). In our experiments, we used a split of 75 subjects (150 scans) for training, 25 subjects (50 scans) for validation. Slices within 3D-MRI scans were considered as 2D images, themselves randomly cropped into patches of size 192$\times$192. These patches are fed as input to the network.

\noindent\textbf{Prostate MR Image Segmentation (P\textsc{romise}12) Challenge:} This dataset comprises multi-centric transversal T2-weighted MR images from 50 subjects. Volumetric images were acquired with multiple MRI vendors and different scanning protocols, and are thus representative of typical MR images acquired in a clinical setting. Image resolution ranges from $15 \times 256 \times 256$ to $54 \times 512 \times 512$ voxels with a spacing ranging from $2 \times 0.27 \times 0.27$ to $4 \times 0.75 \times 0.75$ $\mr{mm}^3$. We slice volumetric images to 2D images along short-axis, and then randomly crop these images into input patches of size $192 \times 192$. We randomly select 40 subjects as our training set and use the remaining 10 subjects as validation set.

\begin{table}[h]
\centering
\caption{Mean DSC, HD and non-connected pixels for segmenting the left ventricle (LV), myocardium (Myo) of ACDC, and segmenting prostate of P\textsc{romise}12. 
}
\label{tab:additional_results}
\setlength{\tabcolsep}{5pt}
\renewcommand{\arraystretch}{.9}
\begin{scriptsize}
\begin{tabular}{llrrr}
\toprule
\B Task / \B labeled\,\% & \B Method & \B DSC\,(\%)\,$\uparrow$ & \B HD\,(mm)\,$\downarrow$ & \B \,N-conn\,(\%)\,$\downarrow$ \\ 
\midrule
\multirow{9}{*}{ACDC LV / 5\,\%} & Baseline & 88.47 (0.58) & 10.12 (0.80) & 6.31 (0.31)\\
 & Entropy min & 88.89 (0.45) & 11.98 (1.47) & 4.99 (1.06)\\
 & VAT & 89.80 (0.58) & 13.38 (2.95) & 5.31 (0.80)\\
 & Co-training & 91.35 (0.56) & 8.63 (2.93) & 5.44 (0.66)\\
 & Mean Teacher & 91.25 (0.48) & 9.10 (1.27) & \second{3.26} (0.41)\\
\cmidrule(l{5pt}r{5pt}){2-5}
 & \CaVAT\,($r_u=0$) & 89.42 (0.75) & 15.34 (2.57) & 4.80 (0.05)\\
 & \CaVAT & 89.89 (0.37) & 12.15 (2.97) & 4.67 (0.67)\\
 & CoT\,+\,\CaVAT & \first{91.77} (0.27) & \first{6.20} (0.55) & 5.36 (0.34)\\
 & MT\,+\,\CaVAT & \second{91.57} (0.42) & \second{7.54} (0.45) & \first{2.32} (0.63)\\
\specialrule{1pt}{2pt}{2pt}
\multirow{9}{*}{ACDC Myo / 5\,\%} & Baseline &  77.70 (1.22) & 12.33 (0.69) & 9.30 (1.03)\\
 & Entropy min &  77.09 (1.24) & 14.68 (3.61) & 9.31 (0.83)\\
 & VAT &  81.19 (0.68) & 17.36 (0.76) & 5.60 (0.55)\\
 & Co-training & 77.88 (1.19) & 12.94 (3.84) & 10.86 (1.09)\\
 & Mean Teacher & \first{84.28} (0.15) & \second{11.31} (1.77) & \second{3.20} (0.82)\\
\cmidrule(l{5pt}r{5pt}){2-5}
& \CaVAT\,($r_u=0$) & 81.68 (0.86) & 15.15 (4.82) & 5.69 (1.03)\\
 & \CaVAT & 81.80 (0.28) & 19.20 (2.81) & 4.53 (0.25)\\
 & CoT\,+\,\CaVAT & 80.21 (0.41) & 12.30 (2.94) & 8.27 (1.11)\\
 & MT\,+\,\CaVAT & \second{84.26} (0.16) & \first{9.43} (0.62) & \first{1.75} (0.47)\\
\specialrule{1pt}{2pt}{2pt}
\multirow{8}{*}{P\textsc{romise}12 /  8\,\%} & Baseline & 66.79 (2.59) & 9.75 (0.15) & 21.73 (5.37) \\
& Entropy min & 68.68 (0.79) & 8.66 (0.52) & 21.28 (1.29) \\
& VAT & 73.33 (0.64) & 9.87 (0.81) & 13.16 (0.53) \\
& Co-training & 67.64 (0.84) & 8.68 (0.87) & 24.07 (1.49) \\
& Mean Teacher & 75.08 (0.89) & \first{8.48} (1.21) & 16.17 (1.46) \\
\cmidrule(l{5pt}r{5pt}){2-5}
& \CaVAT($r_u=0$) & 73.53 (0.89) & 11.76 (1.02) & 11.74 (0.65)\\
& \CaVAT & 75.37 (1.79) & 11.58 (1.07) & \first{9.44} (0.40) \\
& CoT\,+\,\CaVAT & \second{75.47} (0.67) & \second{8.52} (0.65) & 13.02 (1.36) \\
& MT\,+\,\CaVAT & \first{77.24} (0.48) & 8.65 (1.19) & \second{11.68} (2.07) \\
\bottomrule
\end{tabular}
\end{scriptsize}
\vspace{-0mm}
\end{table}

\begin{table}[h]
\centering
\caption{Impact of constraint loss weight $\gamma$ on the ACDC Myo segmentation task with $5\%$ labeled data.}
\label{tab:Myo_consweight}
\setlength{\tabcolsep}{5pt}
\renewcommand{\arraystretch}{.9}
\begin{scriptsize}
\begin{tabular}{cp{50pt}rr}
\toprule
\multirow[b]{2}{*}{\B Labeled\,\%} & \multirow[b]{2}{*}{\B $\gamma$} & \multicolumn{2}{c}{\B Myo}\\
\cmidrule(l{6pt}r{6pt}){3-4}
 & & \multicolumn{1}{c}{\B DSC\,(\%)\,$\uparrow$} & \multicolumn{1}{c}{\B N-conn \,(\%)\,$\downarrow$}\\
\midrule
\multirow{6}{*}{5\,\%}  
& $\gamma=0.01$ & 79.26 (1.25) & 3.17 (0.99)\\
& $\gamma=0.008$ & 81.24 (0.71) & 3.33 (0.50)\\ 
& $\gamma=0.006$ & 81.61 (0.77) & 5.12 (0.65)\\ 
& $\gamma=0.005$ & 81.80 (0.28) & 4.53 (0.25)\\ 
& $\gamma=0.001$ & 79.30 (1.43) & 8.50 (0.48)\\ 
& $\gamma=0.0005$ & 78.78 (0.96) & 8.22 (0.76)\\ 
& $\gamma=0.0001$ & 78.30 (0.25) & 8.57 (1.08)\\ 
\bottomrule
\end{tabular}
\end{scriptsize}
\vspace{-4mm}
\end{table}

\textbf{Results}\\
\tabref{tab:additional_results} provides additional results on the tasks of segmenting LV and Myo of ACDC with 5\% labeled data, and segmenting prostate of P\textsc{romise}12 with 8\% labeled data. As in the test cases reported in the main paper, we see that our \CaVAT{} method provides a better accuracy and foreground connectivity than the baseline using only labeled images (Baseline). Moreover, when added on top of a Co-training or Mean Teacher, it provides a notable reduction in the number of non-connected foreground pixels (N-conn), while also giving a comparable or even better segmentation accuracy (DSC and HD). \tabref{tab:Myo_consweight} reports the performance of our method for different values of constraint loss weight $\gamma$. It can be observed that increasing $\gamma$ up to $0.005$ improves segmentation performance consistently with an increased DSC and reduced N-conn. On the other hand, using a too large $\gamma$ hurts the performance. This may be due to having a too important adversarial noise which makes the network optimization unstable.

\end{document}